\documentclass[letterpaper]{article} 
\usepackage{arxiv}  
\usepackage[hyphens]{url}  
\usepackage{graphicx} 
\usepackage{natbib}  
\usepackage{caption} 
\usepackage{algorithm}
\usepackage{algorithmic}

\usepackage{newfloat}
\usepackage{listings}
\DeclareCaptionStyle{ruled}{labelfont=normalfont,labelsep=colon,strut=off} 
\floatstyle{ruled}
\newfloat{listing}{tb}{lst}{}
\floatname{listing}{Listing}

\usepackage{booktabs}

\usepackage{amsmath}
\usepackage{graphicx}
\usepackage{amssymb}
\usepackage{bbding}
\usepackage{booktabs}
\usepackage{tabularx}

\usepackage{colortbl}
\usepackage{xcolor}
\definecolor{mygray}{gray}{.9} 
\definecolor{myred}{rgb}{1,0,0} 
\definecolor{mygreen}{rgb}{0,1,0} 
\definecolor{myblue}{rgb}{0,0,1} 
\definecolor{mycyan}{rgb}{0,1,1} 
\definecolor{mymagenta}{rgb}{1,0,1} 
\definecolor{myyellow}{rgb}{1,1,0} 
\definecolor{myorange}{rgb}{1,0.647,0} 

\usepackage{subcaption}
\usepackage{multirow}
\usepackage{diagbox}
\usepackage{bm}
\usepackage{amsthm}

\definecolor{LightRed}{RGB}{255,182,193} 
\definecolor{LightBlue}{RGB}{173,216,230}

\theoremstyle{plain}

\theoremstyle{definition}

\theoremstyle{remark}

\usepackage{xspace}

\title{HistoGPA: A Context-Conditioned Gene-Prior Attention Framework for
Histology-Based Spatial Gene Expression Prediction}

\author{
    Ziang Liu\textsuperscript{\rm 1,2,3},
    Qinglin Wang\textsuperscript{\rm 1,2,3},
    Yigui Feng\textsuperscript{\rm 1,2,3},
    Shuai Li\textsuperscript{\rm 1,2,3},
    Jie Liu\textsuperscript{\rm 1,2,3}\thanks{Corresponding author: Jie Liu (liujie@nudt.edu.cn).}
}
\affiliations{
 
    \textsuperscript{\rm 1}Laboratory of Digitizing Software for Frontier Equipment, National University of Defense Technology\\
    \textsuperscript{\rm 2}National Key Laboratory of Parallel and Distributed Computing, National University of Defense Technology\\
    \textsuperscript{\rm 3}College of Computer Science and Technology, National University of Defense Technology, Changsha 410073, China\\

}

\usepackage{bibentry}

\begin{document}

\maketitle

\begin{abstract}
Spatial transcriptomics is costly, motivating the prediction of spatial gene expression from routine hematoxylin and eosin (H\&E) slides. Existing methods use tissue context to refine visual features but keep pretrained gene priors fixed across specimens. Since the relevance of such a prior may vary with tissue context, we propose HistoGPA, which conditions the gene prior itself on the slide being analyzed. A slide-level histological representation modulates a memory of pretrained gene embeddings before each spatial location queries it through cross-attention, and also gates context into the local features. Across ten cancer-specific tasks from HEST-1k, HistoGPA achieves a macro-averaged gene-wise Pearson correlation of 0.300 on the HEST-Bench genes, compared with 0.260 for the strongest of four recent methods. Capacity-matched ablations demonstrate the importance of slide-specific information and show that conditioning the gene prior provides benefits beyond visual context modulation, while pretrained gene embeddings outperform random tokens.
\end{abstract}

\section{Introduction}
By profiling transcripts at their positions within a tissue section~\cite{visualization}, spatial transcriptomics (ST) allows gene expression to be related to tissue organization, supporting studies that range from tumor heterogeneity and microenvironments~\cite{miller} to spatially resolved cell--cell communication~\cite{cang2023commot}.
However, large-scale ST profiling remains constrained by cost and
platform-specific experimental requirements~\cite{cipurko2025arrayseq}.
By comparison, hematoxylin and eosin (H\&E) slides are prepared routinely in clinical pathology, and their morphology reflects aspects of the underlying molecular state of the tissue~\cite{schmauch2020he2rna}. Inferring spatially resolved expression directly from such images, referred to as histology-based spatial gene expression prediction, therefore offers a way to obtain spatial molecular information at the scale of routine pathology.

Despite recent progress, histology-to-expression prediction remains
inherently context dependent.
At the visual level, the molecular interpretation of a local tissue region
may depend on the broader histological context of the specimen.
Morphologically similar regions can occur in different tissue compartments
or pathological environments yet exhibit distinct transcriptional patterns.

A related issue concerns how pretrained molecular knowledge is used for
histology-based expression prediction.
Pretrained gene embeddings are learned from large and heterogeneous
transcriptomic corpora~\cite{geneformer,scgpt} and encode transferable
relationships among genes.
By design, these representations are shared across samples and provide a
generic molecular prior.
However, when such a prior is used to interpret the morphology of a
particular specimen, the relevance of its molecular features may depend on
the specimen-specific tissue context.
We therefore hypothesize that histological context can play two complementary roles: refining local visual representations and adapting the molecular prior before it interacts with local morphology.

Existing approaches address these two sources of information from complementary directions. Spatial, multi-scale, and whole-slide methods enrich local visual representations with broader tissue context~\cite{triplex,hist2st,stflow,dkan}, while cross-modal and gene-aware methods incorporate molecular information through representation alignment or pretrained gene representations~\cite{hage,bleep,dkan}. These studies establish the value of both tissue context and molecular knowledge. However, the pretrained gene representations in these methods are shared across specimens before they interact with image features, so slide context can affect them only indirectly, through the visual features they are matched with. This leaves open whether the slide-level context of the current specimen can be used to condition the molecular prior itself, in addition to refining visual representations.

To address this gap, we propose HistoGPA, a context-conditioned framework for histology-based spatial gene expression prediction. The key idea is to use a shared slide-level histological representation to jointly adapt local morphology and a pretrained molecular prior before prediction. Specifically, Global Context Modulation (GCM) refines position-aware morphological features according to slide context, while Gene-Prior Attention (GPA) uses the same context to condition a pretrained gene-embedding memory before morphology-guided retrieval. The resulting representations are fused for joint multi-gene prediction at each spatial location. Across ten independently trained cancer-specific tasks in HEST-1k~\cite{hest}, HistoGPA achieves a macro-averaged gene-wise PCC of 0.300, compared with 0.260 for the strongest compared method. Ablations further show that most of this improvement depends on slide-specific information, that conditioning the gene prior remains beneficial when the local features already receive slide context, and that the pretrained gene embeddings outperform randomly initialized tokens.
\section{Related Work}

\subsection{Histology-Based Spatial Gene Expression Prediction}

Histology-based spatial gene expression prediction has been studied through
spatial-context modeling, cross-modal representation learning, gene-aware
modeling, and generative approaches.
Spatial-context methods enrich local morphology with information from
surrounding tissue.
Hist2ST~\cite{hist2st} models spatial dependencies using transformer and
graph-based architectures, while TRIPLEX~\cite{triplex} integrates
multi-resolution visual features to capture tissue context across spatial
scales.
Cross-modal approaches instead exploit relationships between histological
and molecular representations.
BLEEP~\cite{bleep} aligns image patches with expression profiles through
bimodal contrastive learning.
HistoPrism~\cite{histoprism}, in a complementary direction, extends
histology-based expression prediction to a pan-cancer setting with
pathway-level evaluation.

Gene-aware methods more directly incorporate external molecular knowledge.
HAGE~\cite{hage} introduces Gene2Vec embeddings encoding gene co-expression
relationships into pathology representations through cross-attention and
hierarchical image--expression alignment.
DKAN~\cite{dkan} encodes gene knowledge summarized by a large language model with a biomedical text encoder and uses the resulting gene-semantic features as queries over multi-scale image features, including whole-slide context, within a dual-path contrastive alignment framework. These studies establish the utility of incorporating gene knowledge into histology-based prediction. In both methods, however, the gene representations are shared across specimens before they interact with image features.
HistoGPA does not seek novelty from introducing gene information itself;
instead, it focuses on specimen-specific contextual adaptation, using the
slide-level histological representation of the current specimen to condition
a pretrained gene-embedding memory before morphology-guided retrieval and,
in parallel, to modulate local morphological features.

Generative approaches provide a complementary direction.
Stem~\cite{stem} uses diffusion-based modeling for spatial expression
generation, while STFlow~\cite{stflow} models joint whole-slide expression
profiles through flow matching.
These methods primarily focus on modeling the conditional distribution of
expression, whereas HistoGPA follows a deterministic prediction setting and
focuses on context-conditioned integration of histological and pretrained
molecular representations.

\subsection{Foundation Models for Histology and Transcriptomics}

Foundation models provide complementary pretrained representations for
histological and molecular data.
Pathology foundation models such as UNI~\cite{uni}, GigaPath~\cite{gigapath},
and CONCH~\cite{conch} learn transferable visual representations from
large-scale histology data, with GigaPath additionally supporting
whole-slide contextual aggregation.
Transcriptomic foundation models, including
Geneformer~\cite{geneformer}, scGPT~\cite{scgpt},
scFoundation~\cite{scfoundation}, CellFM~\cite{cellfm}, and
Nicheformer~\cite{nicheformer}, learn transferable representations of genes
and cellular states from large transcriptomic corpora.
Rather than developing a new foundation model, HistoGPA investigates how
pretrained histological and molecular representations can be integrated
through specimen-specific contextual adaptation, using slide-level
histological context to condition both local morphology and a pretrained
molecular prior.
\section{Method}
\begin{figure*}[t]
    \centering
    \includegraphics[width=0.98\textwidth]{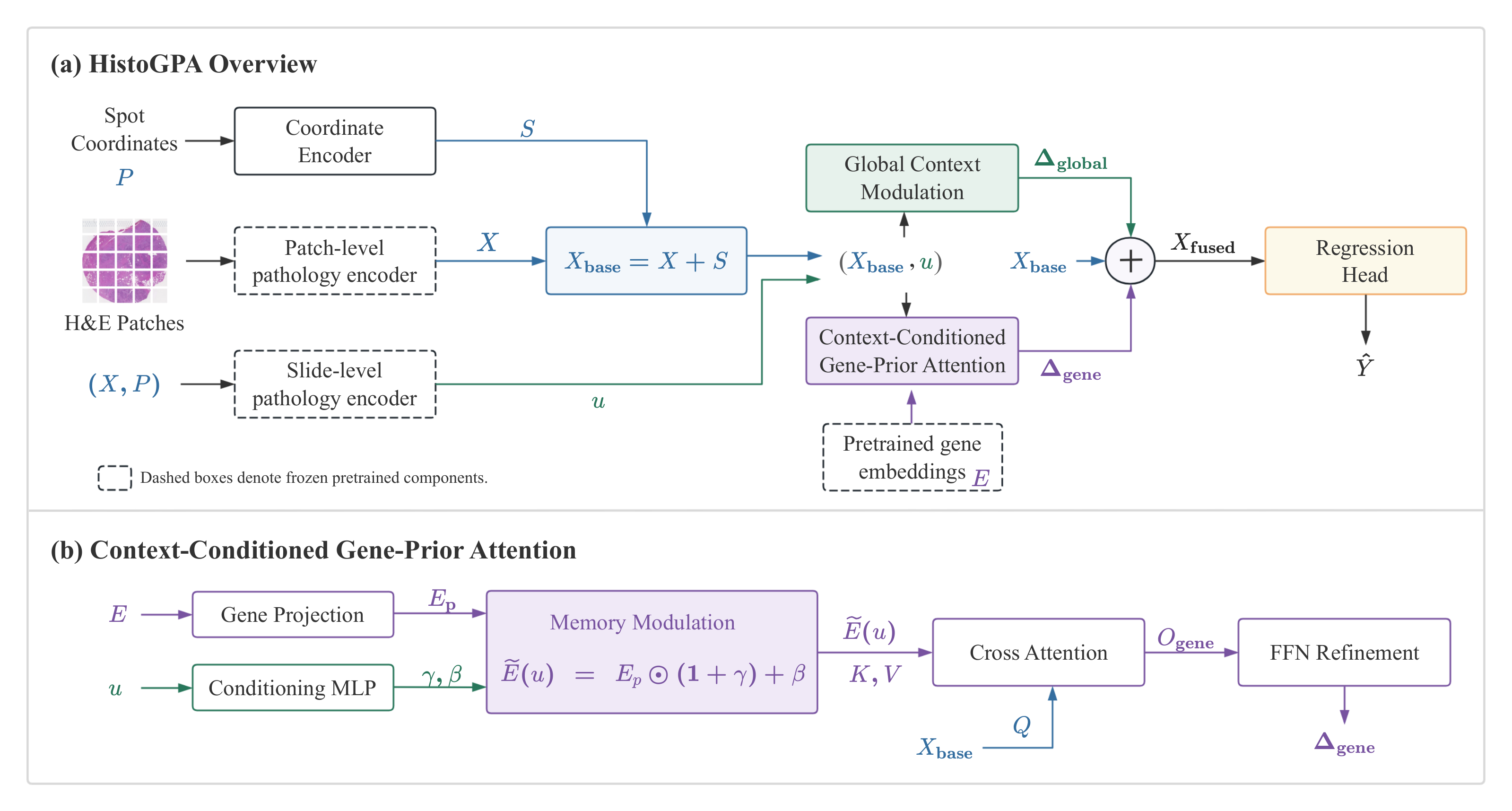}
    \caption{
    \textbf{Overview of HistoGPA.}
    (a) A shared slide-level histological representation conditions two
    complementary pathways.
    GCM injects slide context into the position-aware local representation,
    while GPA uses the same context to adapt a pretrained gene-embedding
    memory before morphology-guided retrieval.
    Their residuals are fused with the position-aware base representation
    for joint prediction of the $G$ target genes.
    (b) GPA adapts the projected gene memory using slide context and
    retrieves from the conditioned memory with $X_{\mathrm{base}}$ as the
    query.
    Dashed boxes denote frozen pretrained components.
    }
    \label{fig:framework}
\end{figure*}

\subsection{Problem Formulation}

We consider histology-based spatial gene expression prediction over a
fixed target gene panel.
Given an H\&E-stained whole-slide image, let
$X=[x_1,\ldots,x_N]^\top\in\mathbb{R}^{N\times d}$ denote the
patch-level histological representations associated with $N$ assayed
spatial locations, where $x_i$ corresponds to the image patch centered
at location $i$.
The spatial coordinates of these locations in the histology image
coordinate system are denoted by
$P \in \mathbb{R}^{N\times2}$.
In addition to the local representations, we use a slide-level
histological representation
$u \in \mathbb{R}^{d_u}$
to summarize shared histological context across the assayed locations
of the current slide.

For a target panel containing $G$ genes, we use pretrained gene
embeddings
$E=[e_1,\ldots,e_G]^\top\in\mathbb{R}^{G\times d_{\mathrm{gene}}}$,
where $e_j$ is retrieved according to the identity of gene $j$.
The embedding set is treated collectively as a pretrained molecular prior
memory from which location-specific information can be retrieved.

The measured spatial expression matrix is denoted by
$Y=[y_1,\ldots,y_N]^\top\in\mathbb{R}^{N\times G}$, where $y_i$
is the preprocessed expression profile at spatial location $i$ and
$Y_{ij}$ is the corresponding expression value of gene $j$.
Our goal is to learn
\begin{equation}
    \widehat{Y}
    =
    f_{\theta}(X,P,u,E)
    \in\mathbb{R}^{N\times G},
    \label{eq:problem}
\end{equation}
where each row $\widehat{y}_i$ represents the predicted expression
profile at location $i$.

\subsection{Overview of HistoGPA}

As illustrated in Figure~\ref{fig:framework}, HistoGPA takes the local
histological representation $X$, slide-level context $u$, spatial
coordinates $P$, and pretrained gene embeddings $E$ defined above and
integrates them through two context-conditioned pathways.
The histological representations are extracted using pretrained GigaPath~\cite{gigapath}
tile and slide encoders, while the gene embeddings are retrieved from
scGPT~\cite{scgpt} according to the identities of the target genes.
These pretrained components remain frozen during HistoGPA training.

We first augment the local histological representation with explicit
spatial information.
After normalizing the spot coordinates while preserving the relative
tissue geometry, we encode them as
\begin{equation}
    S
    =
    \phi_{\mathrm{coord}}
    \left(
        \mathrm{PE}_{2D}(\bar{P})
    \right)
    \in\mathbb{R}^{N\times d},
\end{equation}
where $\bar{P}$ denotes the normalized coordinates.
The spatial encoding is added to the local representation to obtain the
position-aware base representation
\begin{equation}
    X_{\mathrm{base}}
    =
    X + S
    \in\mathbb{R}^{N\times d}.
    \label{eq:base_representation}
\end{equation}

Starting from $X_{\mathrm{base}}$, HistoGPA uses the shared slide-level
representation $u$ to condition two complementary pathways.
Global Context Modulation (GCM) injects specimen-level histological context
into the local representation and produces a contextual residual
$\Delta_{\mathrm{global}}$.
In parallel,  Gene-Prior Attention (GPA) uses the same
slide context to adapt the pretrained molecular prior $E$ before
$X_{\mathrm{base}}$ retrieves location-specific information from the
conditioned memory, yielding a molecular-prior residual
$\Delta_{\mathrm{gene}}$.
The two residuals are subsequently fused with $X_{\mathrm{base}}$ and
mapped to the predicted expression profiles $\widehat{Y}$.
We next describe the two context-conditioned pathways in detail, beginning
with GCM.

\subsection{Global Context Modulation}

Global Context Modulation (GCM) injects slide-level histological context
into each spatial location through location-dependent feature-wise
gating.
We first broadcast the slide representation $u$ over the $N$ spatial
locations,
\begin{equation}
    U
    =
    \mathbf{1}_N u^\top
    \in\mathbb{R}^{N\times d_u},
\end{equation}
and compute a modulation gate from the position-aware patch
representation and the shared slide context:
\begin{equation}
    M_{\mathrm{global}}
    =
    \sigma
    \left(
        \operatorname{Linear}_{g}
        \left([X_{\mathrm{base}}\mathbin{\|}U]\right)
    \right),
    \label{eq:gcm_gate}
\end{equation}
where $\operatorname{Linear}_{g}$ is a learned affine projection from
$d+d_u$ to $d$ dimensions, $\sigma$ denotes the element-wise sigmoid
function, $\mathbin{\|}$ denotes feature concatenation, and
$M_{\mathrm{global}}\in\mathbb{R}^{N\times d}$.

In parallel, the slide representation is projected into the patch-feature
space:
\begin{equation}
    U_{\mathrm{global}}
    =
    \phi_{\mathrm{global}}(U)
    \in\mathbb{R}^{N\times d},
    \label{eq:gcm_projection}
\end{equation}
where $\phi_{\mathrm{global}}$ consists of a linear projection followed
by layer normalization and GELU activation.
The resulting contextual residual is
\begin{equation}
    \Delta_{\mathrm{global}}
    =
    M_{\mathrm{global}}
    \odot
    U_{\mathrm{global}},
    \label{eq:gcm}
\end{equation}
where $\odot$ denotes element-wise multiplication.

Although $U_{\mathrm{global}}$ is shared across spatial locations, the
gate $M_{\mathrm{global}}$ depends on each local representation.
Consequently, different locations can select different feature
dimensions from the same slide-level context.
GCM itself does not directly exchange features between spatial
locations; instead, information aggregated across the assayed
locations is provided through the shared GigaPath slide representation
$u$ and injected according to local morphology and position.

\subsection{Context-Conditioned Gene-Prior Attention}

Gene-Prior Attention (GPA) treats the pretrained gene-embedding set as a
molecular prior memory whose representation is adapted to the current
slide before retrieval.
We first project the gene embeddings into the same feature space as the
histological representations:
\begin{equation}
    E_p
    =
    \phi_{\mathrm{gene}}(E)
    \in\mathbb{R}^{G\times d},
    \label{eq:gene_projection}
\end{equation}
where $\phi_{\mathrm{gene}}$ applies a shared linear projection followed
by layer normalization to every gene embedding.

To adapt the memory to the current histological context, the slide
representation $u$ is mapped to two feature-wise modulation vectors:
\begin{equation}
    [\gamma,\beta]
    =
    \phi_{\mathrm{cond}}(u),
    \qquad
    \gamma,\beta\in\mathbb{R}^{d},
    \label{eq:gene_condition}
\end{equation}
where $\phi_{\mathrm{cond}}$ is a two-layer MLP with GELU activation.
The projected molecular memory is then conditioned as
\begin{equation}
    \widetilde{E}(u)
    =
    E_p\odot(1+\gamma)+\beta.
    \label{eq:conditioned_memory}
\end{equation}

This feature-wise affine modulation follows FiLM~\cite{perez2018film}; here it is applied to the gene-embedding memory rather than to visual features.
The modulation vectors $\gamma$ and $\beta$ are slide specific but shared
across all target genes.
Thus, the slide context adapts the feature space of the molecular memory,
while gene-dependent variation remains encoded by the pretrained
representations in $E_p$.
The residual parameterization $1+\gamma$ preserves the identity
transformation when $\gamma=\beta=0$.

The position-aware morphological representations subsequently query the
conditioned memory.
We define
\begin{equation}
\begin{aligned}
    Q_{\mathrm{gene}} &= \operatorname{LN}_{q}(X_{\mathrm{base}}),\\
    K_{\mathrm{gene}} = V_{\mathrm{gene}} &= \operatorname{LN}_{kv}\bigl(\widetilde{E}(u)\bigr),
\end{aligned}
\end{equation}
and compute
\begin{equation}
    O_{\mathrm{gene}}
    =
    \operatorname{MHA}
    \left(
        Q_{\mathrm{gene}},
        K_{\mathrm{gene}},
        V_{\mathrm{gene}}
    \right)
    \in\mathbb{R}^{N\times d}.
    \label{eq:gpa}
\end{equation}

For each spatial location, $O_{\mathrm{gene}}$ aggregates information
over the conditioned gene-embedding set according to its morphological
and positional representation, yielding a spot-level molecular-prior
representation.
Importantly, GPA is designed as a set-valued prior memory rather than a
gene-indexed decoder.
Accordingly, the ordering of memory entries does not define a
correspondence between individual gene tokens and output channels.
The pretrained embeddings contribute through the information encoded in their feature representations, whereas gene-specific output channels are learned subsequently by the supervised regression head.

We further refine the retrieved molecular representation using a
pre-norm feed-forward block whose input combines the base representation
and the cross-attention output:
\begin{equation}
    \Delta_{\mathrm{gene}}
    =
    O_{\mathrm{gene}}
    +
    \operatorname{FFN}
    \left(
        \operatorname{LN}_{\mathrm{ffn}}
        \left(
            X_{\mathrm{base}} + O_{\mathrm{gene}}
        \right)
    \right).
    \label{eq:gpa_ffn}
\end{equation}
The resulting
$\Delta_{\mathrm{gene}}\in\mathbb{R}^{N\times d}$ represents the
molecular-prior residual at each spatial location.

\subsection{Prediction and Training Objective}

The two context-conditioned residuals are combined with the
position-aware base representation:
\begin{equation}
    X_{\mathrm{fused}}
    =
    X_{\mathrm{base}}
    +
    \Delta_{\mathrm{global}}
    +
    \Delta_{\mathrm{gene}}.
    \label{eq:fusion}
\end{equation}
After layer normalization, a projection module $\phi_{\mathrm{proj}}$
maps the fused features to latent spatial representations
\begin{equation}
    H
    =
    \phi_{\mathrm{proj}}
    \left(\operatorname{LN}_{\mathrm{final}}(X_{\mathrm{fused}})\right)
    \in\mathbb{R}^{N\times d_h},
\end{equation}
and an MLP regressor $\phi_{\mathrm{reg}}$ jointly predicts the $G$
target genes.
Since the preprocessed expression targets are non-negative, we constrain
the predictions as
\begin{equation}
    \widehat{Y}
    =
    \operatorname{ReLU}
    \left(
        \phi_{\mathrm{reg}}(H)
    \right)
    \in\mathbb{R}^{N\times G}.
    \label{eq:prediction}
\end{equation}
The output channels follow the fixed target-gene order and acquire their
gene-specific roles through supervised training rather than through
memory-row indices.

We train HistoGPA using the standard mean-squared error (MSE) objective:
\begin{equation}
    \mathcal{L}_{\mathrm{MSE}}
    =
    \frac{1}{|\Omega|}
    \sum_{(i,j)\in\Omega}
    \left(
        \widehat{Y}_{ij}-Y_{ij}
    \right)^2,
    \label{eq:mse_loss}
\end{equation}
where $\Omega$ denotes the set of valid location--gene pairs used for supervision. All trainable components of HistoGPA are optimized end-to-end using this objective.

\section{Experiments}
\begin{table*}[t]
\centering
\caption{
Gene-wise PCC on the predefined 50-gene HEST-Bench subset.
Cancer-type results are mean $\pm$ SD across 20 independent runs;
Macro Avg.\ is the unweighted mean of the ten cancer-type mean PCCs, reported as mean $\pm$ SD of the per-run macro averages.
Bold indicates the highest mean in each row.
}
\label{tab:main_results}

\footnotesize
\setlength{\tabcolsep}{3.5pt}
\renewcommand{\arraystretch}{1.10}

\newcommand{\msd}[2]{\ensuremath{#1\,{\scriptstyle\pm #2}}}
\newcommand{\bmsd}[2]{\ensuremath{\mathbf{#1}\,{\scriptstyle\pm #2}}}

\begin{tabular*}{\textwidth}{@{\extracolsep{\fill}}lccccc@{}}
\toprule
Cancer Type
& STFlow
& HAGE
& HistoPrism
& Stem
& HistoGPA
\\
\midrule
CCRCC
& \msd{0.140}{0.031}
& \msd{0.126}{0.003}
& \bmsd{0.161}{0.040}
& \msd{0.133}{0.017}
& \msd{0.144}{0.004}
\\
COAD
& \msd{0.259}{0.054}
& \msd{0.267}{0.015}
& \msd{0.152}{0.080}
& \msd{0.231}{0.022}
& \bmsd{0.332}{0.020}
\\
HCC
& \msd{0.093}{0.030}
& \msd{0.082}{0.017}
& \msd{0.036}{0.024}
& \msd{0.062}{0.031}
& \bmsd{0.137}{0.014}
\\
IDC
& \msd{0.415}{0.032}
& \msd{0.433}{0.023}
& \msd{0.367}{0.037}
& \msd{0.439}{0.063}
& \bmsd{0.481}{0.015}
\\
LUNG
& \msd{0.340}{0.055}
& \msd{0.289}{0.021}
& \msd{0.390}{0.070}
& \msd{0.374}{0.042}
& \bmsd{0.429}{0.030}
\\
LYMPH-IDC
& \bmsd{0.241}{0.014}
& \msd{0.168}{0.100}
& \msd{0.235}{0.009}
& \msd{0.216}{0.030}
& \msd{0.228}{0.018}
\\
PAAD
& \msd{0.307}{0.062}
& \msd{0.256}{0.025}
& \msd{0.223}{0.088}
& \msd{0.277}{0.054}
& \bmsd{0.411}{0.020}
\\
PRAD
& \msd{0.242}{0.039}
& \msd{0.145}{0.017}
& \bmsd{0.255}{0.052}
& \msd{0.229}{0.082}
& \msd{0.194}{0.007}
\\
READ
& \bmsd{0.256}{0.010}
& \msd{0.195}{0.012}
& \msd{0.147}{0.085}
& \msd{0.212}{0.019}
& \msd{0.206}{0.017}
\\
SKCM
& \msd{0.307}{0.018}
& \msd{0.393}{0.008}
& \msd{0.394}{0.039}
& \msd{0.352}{0.059}
& \bmsd{0.437}{0.019}
\\
\midrule
Macro Avg. & \msd{0.260}{0.022} & \msd{0.235}{0.007} & \msd{0.236}{0.031} & \msd{0.253}{0.024} & \bmsd{0.300}{0.006} 
\\
\bottomrule
\end{tabular*}
\end{table*}
\subsection{Experimental Setup}

\textbf{Dataset and splits.}
We evaluate HistoGPA on HEST-1k~\cite{hest} using the two predefined slide-level split configurations and the corresponding held-out test sets adopted by HistoPrism~\cite{histoprism}. A separate model is trained for each of the ten cancer types using only the training and validation slides of that cancer type, with no parameter sharing across cancer-specific tasks. The numbers of held-out test slides for each cancer type and split are listed  in Appendix~\ref{sup:splits}. We compare HistoGPA with STFlow~\cite{stflow}, HAGE~\cite{hage}, HistoPrism~\cite{histoprism}, and Stem~\cite{stem}.

\noindent\textbf{Comparison protocol.}
All methods use the same frozen GigaPath tile-encoder features
extracted from identical spot-aligned H\&E patches as their
patch-level histological inputs.
HistoGPA additionally uses a slide-level representation computed by the frozen GigaPath slide encoder from the same patch-level features and coordinates. Using this representation to condition the gene prior is part of the proposed method, and Sec.~\ref{sec:ablation} isolates its contribution with capacity-matched ablations.
For the four competing methods, STFlow~\cite{stflow},
HAGE~\cite{hage}, HistoPrism~\cite{histoprism},
and Stem~\cite{stem}, we use their official implementations
and follow their respective official training configurations,
including method-specific architectures, optimization settings,
and training hyperparameters, without additional
method-specific tuning.
All methods are evaluated using the same data partitions,
expression preprocessing, target-gene panels, and
evaluation protocol.

\noindent\textbf{Target-gene panels and preprocessing.}
For each cancer type and split, the prediction task uses a
fixed panel of $G=1{,}500$ target genes.
The panel contains the predefined 50-gene HEST-Bench subset
~\cite{hest} and 1,450 additional genes selected exclusively
from the corresponding training set by ranking genes according
to their expression variance.
The 50 benchmark genes follow the released cancer-type-specific
lists and remain fixed across slides and split configurations,
whereas selection of the remaining genes uses no validation
or test-set information.
Within each split, the resulting 1,500-gene panel is shared
across training, validation, testing, and all compared methods.
All models predict the complete panel, while HistoGPA additionally
uses the corresponding $1{,}500$ pretrained gene embeddings
as its molecular-prior memory.
Expression counts are normalized independently at each spatial
location over the selected target-gene panel to a total of
$10^4$, followed by a log1p transformation
(see Appendix~\ref{sup:preprocessing} for the exact
formulation).

\noindent\textbf{Implementation details.}
We use pretrained GigaPath~\cite{gigapath} tile and slide
encoders as fixed histological feature extractors and
pretrained 512-dimensional scGPT~\cite{scgpt} embeddings
as the fixed molecular prior.
HistoGPA operates in a 1,536-dimensional feature space
and employs a single eight-head cross-attention layer
in GPA.
The trainable components are optimized using AdamW with
a learning rate of $1\times10^{-5}$, a weight decay of
$0.05$, and a batch size of eight slides for at most
1,000 epochs.
For HistoGPA, the checkpoint with the lowest validation
loss is retained within each independent run.
For each method, cancer-specific task, and predefined split,
we conduct 20 independent runs using the same set of
random seeds (1--20).
Further architectural and implementation details are
provided in Appendix~\ref{sup:implementation}.

\noindent\textbf{Evaluation protocol.}
For the main comparisons and ablation studies, gene-wise Pearson
correlation coefficient (PCC) is computed across valid spatial
locations separately for each of the 50 HEST-Bench genes on each
held-out test slide.
For each random seed, gene-wise PCC values are first
averaged over the 50 benchmark genes within each held-out
slide, and then averaged across all held-out test slides
from the two predefined split configurations to obtain
one cancer-level PCC.
We report the mean and standard deviation (SD) of these
cancer-level PCCs across the 20 random seeds.
The reported Macro Avg. is the unweighted mean of the ten cancer-type mean PCCs, and its SD is computed across the per-run macro averages of the 20 random seeds.

\subsection{Comparison with Existing Methods}

Table~\ref{tab:main_results} reports gene-wise PCC on the predefined 50-gene
HEST-Bench subset across ten independently trained
cancer-specific tasks.
HistoGPA achieves the highest macro-averaged PCC of
0.300, outperforming the strongest competing method,
STFlow (0.260), by 0.040.
It also surpasses Stem (0.253), HAGE (0.235), and
HistoPrism (0.236) in terms of macro-averaged PCC.
At the cancer-type level, HistoGPA achieves the highest
mean PCC on six of the ten tasks: COAD, HCC, IDC, LUNG,
PAAD, and SKCM.
The largest improvements over the strongest competing
methods are observed on PAAD (0.411 vs. 0.307) and
COAD (0.332 vs. 0.267).
These results indicate that the aggregate improvement
is supported by gains across multiple cancer-specific
tasks rather than being driven by a single task.

Performance nevertheless varies across cancer types.
HistoGPA does not achieve the highest mean PCC on
CCRCC, LYMPH-IDC, PRAD, or READ.
The differences are relatively small on CCRCC (0.017) and LYMPH-IDC (0.013), whereas larger gaps remain on PRAD (0.061) and READ (0.050).
Overall, these results demonstrate improved aggregate
prediction performance across the evaluated
cancer-specific tasks, while highlighting differences
in performance across cancer types.

\subsection{Ablation Study}
\label{sec:ablation}

Table~\ref{tab:ablation} summarizes four groups of controls: module ablations, gene-prior controls, and two types of slide-context controls. For the latter, $\bar{u}$ denotes the mean of $u$ over the training slides of the corresponding model; it carries the average slide-level context of the training set but no information about the current slide. HistoGPA has the highest mean PCC on each of the ten cancer types against every variant, and per-cancer results are given in Tables~\ref{sup:tab_ablation_main} and~\ref{sup:tab_ablation_context} in the appendix.

\paragraph{Contributions of GCM and GPA.}
Starting from Base, which removes both pathways, retaining GCM alone increases the macro-averaged PCC from 0.226 to 0.250, and retaining GPA alone increases it to 0.256. Combining the two pathways raises it to 0.300, corresponding to gains of 0.050 over w/o GPA and 0.044 over w/o GCM, so the two pathways contribute complementary improvements to overall prediction performance. Since w/o GPA receives the same inputs as HistoGPA, including $u$, and fuses $u$ directly into the local features, this gap also shows that supplying the slide representation to a feature-level predictor does not by itself reach the performance of HistoGPA.

\paragraph{Slide-dependent gene-prior conditioning.}
Without GCM, adding a fixed gene memory (Static) barely changes the macro-averaged PCC relative to Base (0.230 vs.\ 0.226), whereas conditioning the memory on $u$ (w/o GCM) raises it to 0.256 and yields a higher mean PCC on nine of the ten cancer types, with CCRCC as the only exception. To test whether this benefit persists within the full model, Mean context in GPA keeps the conditioning network and all other components of HistoGPA but conditions the gene memory on $\bar{u}$ while GCM still receives $u$. This variant reaches 0.274 and is below HistoGPA on all ten cancer types. Because it has the same parameters as HistoGPA, the gap of 0.026 reflects slide-specific conditioning of the gene memory rather than additional capacity, and it remains when GCM already provides slide-specific context.

\begin{table}[t]
\centering
\caption{Ablation study on the predefined 50-gene HEST-Bench subset, reported as macro-averaged gene-wise PCC over the ten cancer types (mean $\pm$ SD of the per-run macro averages across 20 runs). GCM and GPA give the slide context received by each pathway: $u$, the representation of the current slide; $\bar{u}$, the mean of $u$ over the training slides of the corresponding model; Fixed, an unconditioned gene memory; --, pathway removed. Memory gives the gene tokens used by GPA. Mean-context variants have the same architecture and number of parameters as HistoGPA and are trained independently. $^\dagger$Trained HistoGPA models evaluated with $\bar{u}$ in both pathways, without retraining. $\Delta$ is relative to HistoGPA, which has the highest mean PCC on each of the ten cancer types against every variant. Per-cancer results are given in Tables~\ref{sup:tab_ablation_main} and~\ref{sup:tab_ablation_context} in the appendix.}
\label{tab:ablation}
\footnotesize
\setlength{\tabcolsep}{3pt}
\newcommand{\msd}[2]{\ensuremath{#1\,{\scriptstyle\pm #2}}}
\newcommand{\bmsd}[2]{\ensuremath{\mathbf{#1}\,{\scriptstyle\pm #2}}}
\begin{tabular}{@{}lccccc@{}}
\toprule
Variant & GCM & GPA & Memory & PCC & $\Delta$ \\
\midrule
\multicolumn{6}{@{}l}{\textit{Module ablation}} \\
Base & -- & -- & -- & \msd{0.226}{0.008} & $-0.074$ \\
w/o GPA & $u$ & -- & -- & \msd{0.250}{0.014} & $-0.050$ \\
w/o GCM & -- & $u$ & scGPT & \msd{0.256}{0.008} & $-0.044$ \\
\midrule
\multicolumn{6}{@{}l}{\textit{Gene-prior controls}} \\
Static & -- & Fixed & scGPT & \msd{0.230}{0.012} & $-0.070$ \\
Frozen random & $u$ & $u$ & Random & \msd{0.279}{0.011} & $-0.021$ \\
Trainable random & $u$ & $u$ & Random & \msd{0.282}{0.009} & $-0.018$ \\
\midrule
\multicolumn{6}{@{}l}{\textit{Slide-context controls}} \\
Mean context (both) & $\bar{u}$ & $\bar{u}$ & scGPT & \msd{0.250}{0.008} & $-0.050$ \\
Mean context in GPA & $u$ & $\bar{u}$ & scGPT & \msd{0.274}{0.009} & $-0.026$ \\
Mean context in GCM & $\bar{u}$ & $u$ & scGPT & \msd{0.270}{0.011} & $-0.030$ \\
Test-time mean$^\dagger$ & $\bar{u}$ & $\bar{u}$ & scGPT & \msd{0.271}{0.005} & $-0.029$ \\
\midrule
HistoGPA & $u$ & $u$ & scGPT & \bmsd{0.300}{0.006} & -- \\
\bottomrule
\end{tabular}
\end{table}

\paragraph{Value of slide-specific information.}
Mean context (both) supplies $\bar{u}$ to both pathways and thus retains every component of HistoGPA without any slide-specific information. It reaches 0.250, between Base (0.226) and HistoGPA, and below w/o GCM (0.256), which has fewer components but receives $u$. Most of the improvement of HistoGPA over Base therefore depends on slide-specific information rather than on the additional components alone. The gain from slide-specific context in each pathway is similar whether or not the other pathway also receives it (0.020 to 0.026 for GPA and 0.024 to 0.030 for GCM), indicating approximately additive contributions. We further replace $u$ with $\bar{u}$ in the trained HistoGPA models at test time (Test-time mean). The macro-averaged PCC drops to 0.271 and decreases on all ten cancer types, confirming that the trained model relies on information specific to the current slide. It nevertheless remains above Mean context (both) (0.250), suggesting that slide-specific context during training also improves the learned representations beyond the information used at inference.

\paragraph{Effect of pretrained gene embeddings.}
Replacing the scGPT embeddings with frozen or trainable randomly initialized tokens, while retaining both pathways and slide-dependent memory conditioning, reduces the macro-averaged PCC to 0.279 and 0.282, respectively. Optimizing the random tokens jointly with the model yields only a modest improvement over freezing them, leaving a gap of 0.018 to HistoGPA, so the pretrained scGPT embeddings provide an additional benefit over randomly initialized gene tokens within the same architecture.

\subsection{Spatial Prediction of Cancer-Associated Marker Genes}
\label{sec:marker}

\begin{figure}[t]
    \centering
    \includegraphics[width=\linewidth]{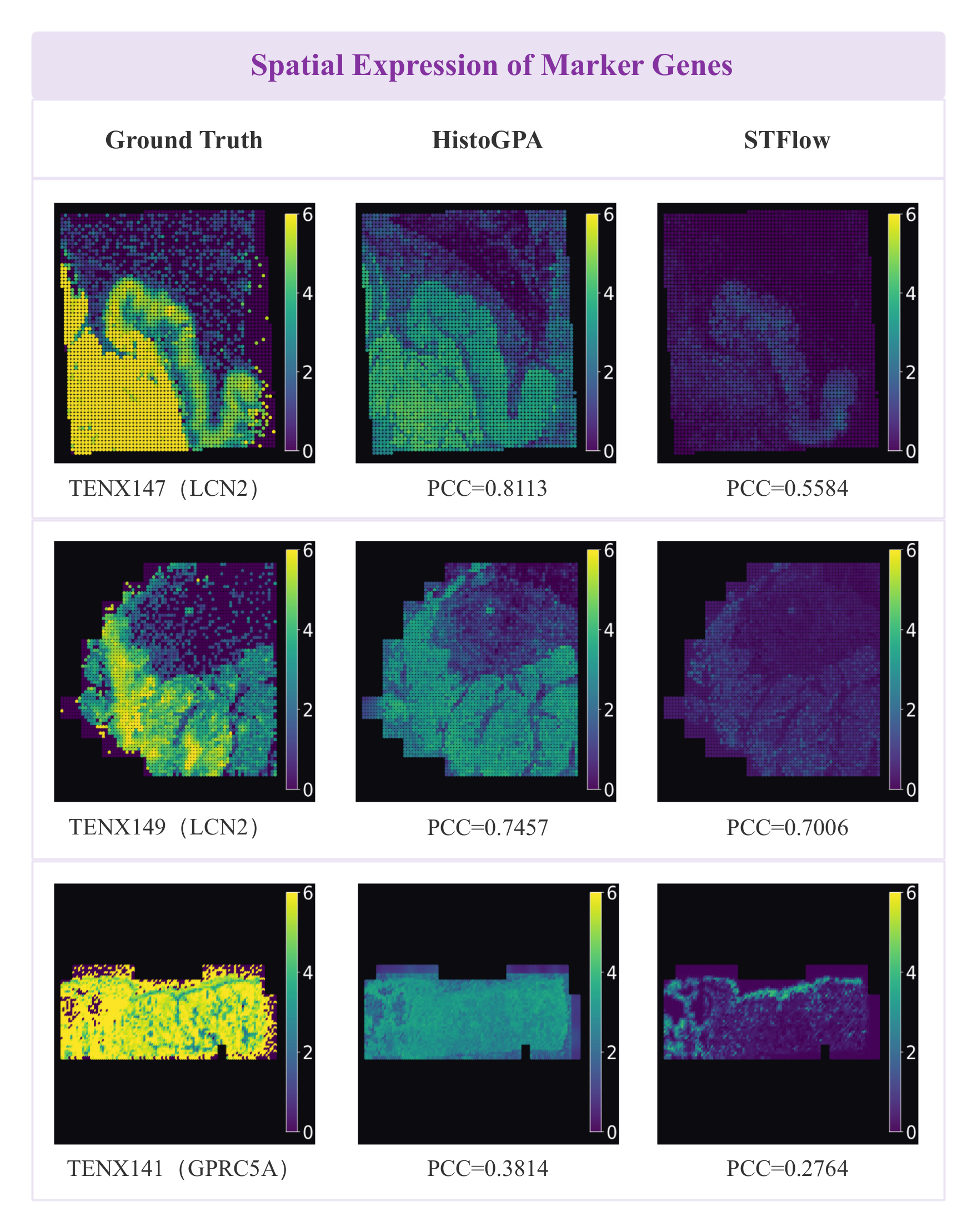}
\caption{
Spatial expression of cancer-associated marker genes on
held-out test slides:
(a) LCN2 on TENX147 (COAD),
(b) LCN2 on TENX149 (COAD), and
(c) GPRC5A on TENX141 (LUNG).
Columns show Ground Truth, HistoGPA, and STFlow, from
left to right.
All methods share the same expression normalization and
0--6 color scale within each row.
PCC across valid tissue locations is reported beneath
each prediction.
Slides are resized independently for visualization while
preserving their aspect ratios. Because PCC is insensitive to the scale of the predictions, brightness differences under the shared color scale mainly reflect prediction magnitude rather than spatial pattern.
}
    \label{fig:marker}
\end{figure}

To complement the aggregate gene-wise PCC evaluation,
we qualitatively examine whether HistoGPA recovers spatial
expression patterns of biologically relevant marker genes.
Similar marker-gene visualizations have also been used in
previous spatial gene expression prediction studies
\cite{stpath,dkan}.
We consider LCN2 in COAD and GPRC5A in LUNG, based on
prior studies linking these genes to the corresponding
cancer types \cite{lcn2,oliveira2025highdefinition,tao2007gprc5a,grpc1}.
Figure~\ref{fig:marker} visualizes the predicted expression patterns on
held-out test slides, comparing HistoGPA with STFlow,
the strongest baseline in terms of macro-averaged PCC
in Table~\ref{tab:main_results}.
The third held-out COAD test slide of split~0, TENX148, is presented in Fig.~\ref{sup:fig_marker} (Appendix~\ref{sup:marker_results}).

For visualization, we independently select the run whose
validation loss is closest to the median across the 20
independent runs for each method, without using test-set
prediction performance for run selection.
The PCC values reported beneath the spatial maps correspond
to these displayed runs.

On the two COAD slides shown in Figure~\ref{fig:marker}, HistoGPA more
closely recovers the measured spatial organization of
LCN2, with a more pronounced difference on TENX147 than
on TENX149.
For GPRC5A on TENX141, both methods produce smoother
expression maps than the measured profile, while HistoGPA
better preserves the dominant large-scale spatial variation.
The additional COAD example in Fig.~\ref{sup:fig_marker} shows a smaller but directionally consistent improvement.
These visualizations provide qualitative evidence of
spatial expression recovery, complementing the aggregate
gene-wise PCC results.

\subsection{Clustering Consistency of Predicted Expression Profiles}

Spatial domain identification is an important downstream
analysis in spatial transcriptomics, as it helps reveal
the organization of distinct tissue regions
\cite{zhu2024multi,xiao2025spatially}.
However, gene-wise PCC evaluates individual targets
independently and does not directly assess whether
predicted expression profiles preserve multigene structure.
We therefore perform clustering consistency analysis
by independently clustering the measured and predicted
expression profiles using the complete 1,500-gene target
panel on each held-out test slide, and evaluating the
agreement between the resulting partitions.

Table~\ref{tab:clustering_metrics} summarizes the macro-averaged clustering consistency across the ten cancer-specific tasks. HistoGPA obtains the highest mean score on all four metrics (ARI 0.281, AMI 0.368, NMI 0.370, homogeneity 0.397), but the margins over the strongest competing methods are small relative to the run-to-run variability (e.g., ARI $0.281\pm0.030$ vs.\ $0.274\pm0.024$ for HistoPrism). We therefore interpret this analysis as showing that the improvement in gene-wise PCC is not obtained at the expense of multigene structure.

\begin{table}[t]
    \centering
\caption{
Macro-averaged clustering consistency across the ten
cancer-specific tasks.
For each independent run, slide-level scores are first
averaged within each cancer-specific task, followed by
an unweighted average across the ten tasks.
Results are reported as mean $\pm$ standard deviation
across 20 independent runs.
Bold indicates the best mean for each metric.
}
    \label{tab:clustering_metrics}

    \renewcommand{\arraystretch}{1.08}
    \setlength{\tabcolsep}{2.5pt}

    \resizebox{\columnwidth}{!}{%
    \begin{tabular}{@{}lcccc@{}}
        \toprule
        Method & ARI & AMI & NMI & Homo. \\
        \midrule
        STFlow
        & $0.221{\pm}0.040$
        & $0.325{\pm}0.036$
        & $0.327{\pm}0.036$
        & $0.367{\pm}0.038$ \\
        Stem
        & $0.254{\pm}0.013$
        & $0.355{\pm}0.010$
        & $0.357{\pm}0.011$
        & $0.351{\pm}0.015$ \\
        HAGE
        & $0.263{\pm}0.013$
        & $0.359{\pm}0.013$
        & $0.361{\pm}0.013$
        & $0.396{\pm}0.013$ \\
        HistoPrism
        & $0.274{\pm}0.024$
        & $0.349{\pm}0.020$
        & $0.350{\pm}0.020$
        & $0.365{\pm}0.024$ \\
        HistoGPA
        & $\mathbf{0.281{\pm}0.030}$
        & $\mathbf{0.368{\pm}0.015}$
        & $\mathbf{0.370{\pm}0.015}$
        & $\mathbf{0.397{\pm}0.011}$ \\
        \bottomrule
    \end{tabular}%
    }
\end{table}

\begin{figure}[t]
    \centering
    \includegraphics[width=\linewidth]{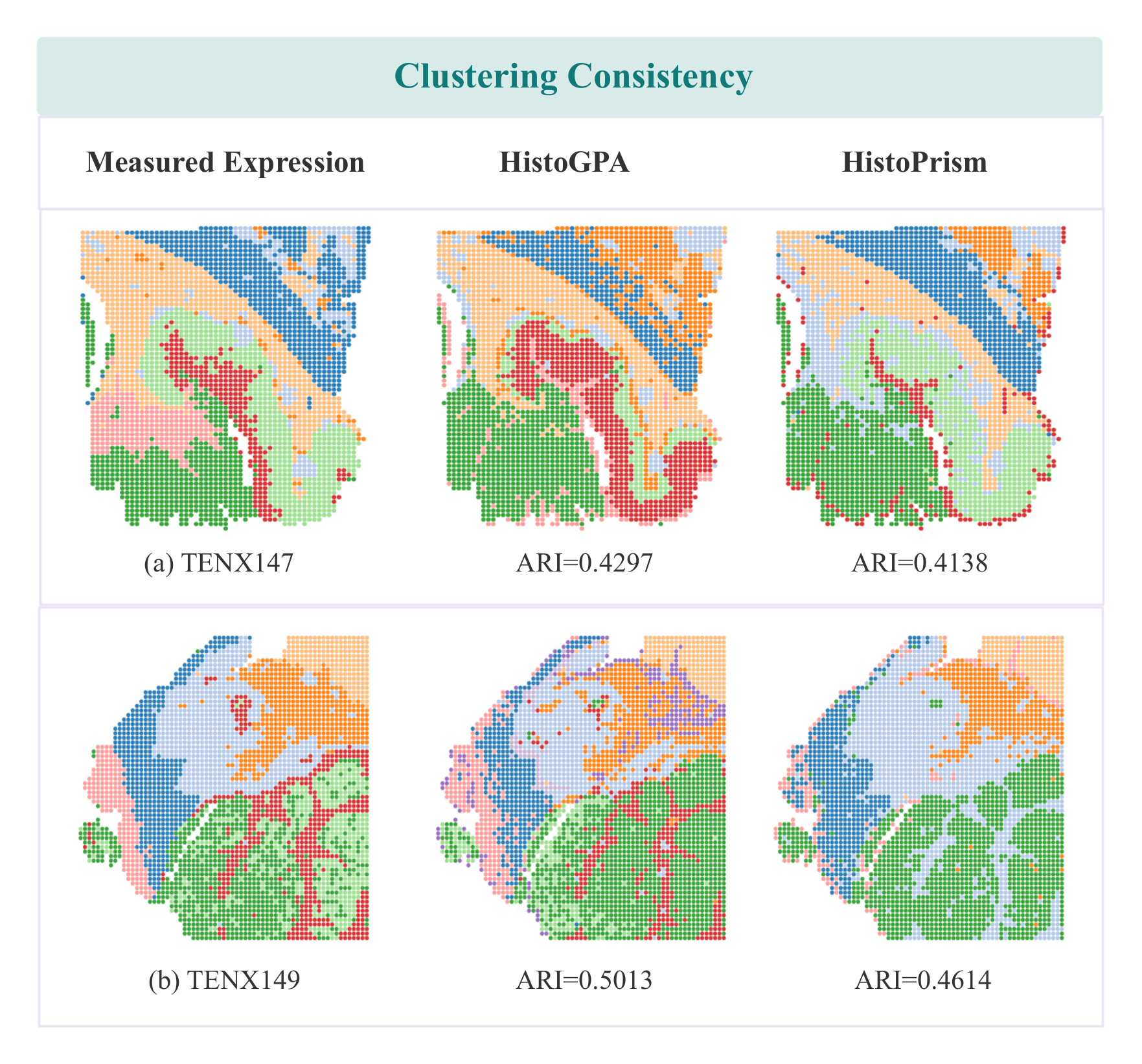}
    \caption{
    Expression-derived clustering on two held-out COAD slides:
    (a) TENX147 and (b) TENX149.
    Columns show partitions independently derived from measured expression,
    HistoGPA predictions, and HistoPrism predictions using the complete
    1,500-gene target panel.
    Displayed predictions use the run whose validation loss is closest to the
    median across the 20 independent runs.
    ARI is reported beneath each prediction.
    Predicted cluster IDs are matched to the measured-expression clusters only
    for visualization after metric computation; colors are comparable within a
    row but not across slides.
    }
    \label{fig:clustering_consistency}
\end{figure}

Figure~\ref{fig:clustering_consistency} shows the resulting partitions on two held-out COAD slides for HistoGPA and HistoPrism, which has the strongest competing macro-averaged ARI, using runs selected by validation loss as in Sec.~\ref{sec:marker}. Both methods recover several major partitions of the measured expression, and HistoGPA attains a higher ARI on both displayed slides.

\section{Conclusion}

We presented HistoGPA, a framework that uses slide-level histological context to condition both local morphological representations and a pretrained gene-embedding memory for spatial gene expression prediction. Across ten cancer-specific tasks in HEST-1k, HistoGPA achieves the highest macro-averaged gene-wise PCC among the compared methods on the predefined 50-gene HEST-Bench subset. Capacity-matched ablations show that most of this improvement depends on slide-specific information and that conditioning the gene memory on the current slide adds to the benefit of slide context in the local features, while pretrained gene embeddings outperform randomly initialized tokens. These findings support adapting molecular priors to the tissue context of each specimen before morphology-guided retrieval.

The current evaluation uses separate cancer-specific models and fixed target-gene panels, and improvements vary across cancer types. The competing methods were evaluated with their original inputs, and whether they would benefit similarly from the GigaPath slide representation was not tested.  Important next steps include extending the framework to shared models across cancer types, evaluating generalization to external datasets, and identifying which properties of the pretrained gene representations account for their predictive benefit.

\bibliography{main}

\appendix

\section{Implementation Details}
\label{sup:implementation}

\subsection{Histological Feature Extraction}
\label{sup:histological_features}

We use the pretrained GigaPath tile and slide encoders as fixed
histological feature extractors.
For each assayed spatial location, we use the
$256\times256$-pixel spot-centered H\&E patch provided with the
HEST-1k dataset and process it using the GigaPath tile encoder.
The resulting local representations form
$X \in \mathbb{R}^{N\times1536}$, where $N$ denotes the number of
assayed locations in the slide.

To construct slide-level context, the same set of spot-level patch
embeddings, together with their spatial coordinates, is further
processed by the pretrained GigaPath slide encoder.
This produces a 768-dimensional slide representation
$u \in \mathbb{R}^{768}$ that summarizes shared histological context
across the assayed locations of the current slide.
Both GigaPath encoders are used only for offline feature extraction
and are not updated during HistoGPA training.

\subsection{Molecular Prior and Model Configuration}
\label{sup:model_configuration}

For each cancer-specific target panel, we retrieve pretrained gene
embeddings by gene identity from the embedding table of the official
pretrained scGPT model.
The resulting molecular prior is
$E \in \mathbb{R}^{1500\times512}$.
These embeddings are precomputed and remain fixed during HistoGPA
training, while the gene projection and all subsequent HistoGPA
components are optimized jointly.

The main architectural dimensions are summarized in
Table~\ref{sup:tab_config}.
The histological feature dimension is $d=1536$.
The 256-dimensional sinusoidal coordinate encoding is mapped to this
feature space by a two-layer coordinate MLP.
Gene embeddings are projected from 512 to 1536 dimensions before
Gene-Prior Attention.
GPA contains a single eight-head cross-attention layer followed by
a feed-forward refinement block.
A dropout rate of $0.1$ is used in the trainable prediction network.

\begin{table}[htbp]
    \centering
    \caption{Main architectural configuration of HistoGPA.}
    \label{sup:tab_config}
    \small
    \begin{tabular}{@{}ll@{}}
        \toprule
        Component & Configuration \\
        \midrule
        Local histological feature $X$ & $1536$ \\
        Slide context $u$ & $768$ \\
        Pretrained gene embedding & $512$ \\
        Coordinate encoding & $256$ \\
        Coordinate MLP & $256 \rightarrow 512 \rightarrow 1536$ \\
        GCM context projection & $768 \rightarrow 1536$ \\
        GCM gating projection & $2304 \rightarrow 1536$ \\
        Gene projection & $512 \rightarrow 1536$ \\
        Conditioning MLP & $768 \rightarrow 1536 \rightarrow 3072$ \\
        GPA cross-attention & $1$ layer, $8$ heads \\
        GPA feed-forward block & $1536 \rightarrow 6144 \rightarrow 1536$ \\
        Feature projector & $1536 \rightarrow 4096 \rightarrow 4096$ \\
        Regression head & $4096 \rightarrow 2048 \rightarrow 1024 \rightarrow 1500$ \\
        Dropout & $0.1$ \\
        \bottomrule
    \end{tabular}
\end{table}

\subsection{Expression Preprocessing and Training Objective}
\label{sup:preprocessing}

Expression values are normalized independently for each spatial
location over the target-gene panel and subsequently log-transformed.
Given the raw count $C_{ij}$ of gene $j$ at location $i$, the
regression target is computed as
\begin{equation}
    Y_{ij}
    =
    \log\left(
        1 +
        10^{4}
        \frac{C_{ij}}
        {\max\left(\sum_{j'=1}^{G} C_{ij'},\,10^{-8}\right)}
    \right).
    \label{sup:eq_preprocessing}
\end{equation}

We train the model using the standard mean-squared error (MSE)
over all valid location--gene pairs:
\begin{equation}
    \mathcal{L}_{\mathrm{MSE}}
    =
    \frac{1}{|\Omega|}
    \sum_{(i,j)\in\Omega}
    \left(\hat{Y}_{ij}-Y_{ij}\right)^2,
    \label{sup:eq_mse}
\end{equation}
where $\Omega$ denotes the set of valid location--gene pairs,
excluding padded locations.

For minibatches containing slides with different numbers of
assayed locations, shorter samples are padded to accommodate
batch processing. The padded locations are excluded from the
training loss, while all target-gene entries at valid assayed
locations contribute equally to the objective.

\subsection{Optimization and Model Selection}
\label{sup:optimization}

HistoGPA is optimized using AdamW with a learning rate of
$1\times10^{-5}$ and a weight decay of $0.05$.
The training batch size is eight slides, with each slide contributing
all available assayed locations used for feature extraction.
Models are trained for at most 1000 epochs, and model selection is
performed exclusively according to validation loss.
The pretrained GigaPath features and scGPT gene embeddings remain
fixed throughout training; all HistoGPA-specific projection,
modulation, attention, feed-forward, and regression components are
trainable.

For each cancer-specific task and predefined data split,
we conduct 20 independent training runs using random seeds
1--20. Reported results are aggregated over these independent
runs according to the evaluation protocol described in
Sec.~4.1.

\subsection{Data Splits}
\label{sup:splits}

We use the two predefined slide-level split configurations
described in Sec.~4.1.
For each cancer-specific task, the model is trained and selected
using only the slides of that cancer type in the training and
validation sets of the corresponding split, and is evaluated on
the held-out test slides of that cancer type
(Table~\ref{sup:tab_splits}).
Six of the ten cancer types have one test slide per split,
so their cancer-level results are based on two held-out slides.
The standard deviations reported in the main text reflect
variation across independent training runs rather than across
test slides.

\begin{table}[htbp]
    \centering
    \caption{Number of held-out test slides for each cancer type
    in the two split configurations.}
    \label{sup:tab_splits}
    \small
    \begin{tabular}{@{}lccc@{}}
        \toprule
        Cancer Type & Split 0 & Split 1 & Total \\
        \midrule
        CCRCC     & 4 & 4  & 8  \\
        COAD      & 3 & 1  & 4  \\
        HCC       & 1 & 1  & 2  \\
        IDC       & 1 & 1  & 2  \\
        LUNG      & 1 & 1  & 2  \\
        LYMPH-IDC & 1 & 1  & 2  \\
        PAAD      & 1 & 1  & 2  \\
        PRAD      & 8 & 15 & 23 \\
        READ      & 2 & 2  & 4  \\
        SKCM      & 1 & 1  & 2  \\
        \midrule
        Total     & 23 & 28 & 51 \\
        \bottomrule
    \end{tabular}
\end{table}

\subsection{Compared Methods}
\label{sup:baselines}

For STFlow, HAGE, HistoPrism, and Stem, we use the official
implementations and follow their official training
configurations, including optimization settings, numbers of
training epochs, and checkpoint selection.
All four methods accept the frozen GigaPath tile features as
patch-level inputs without modification, and they use the same
data partitions, expression preprocessing, target-gene panels,
and evaluation protocol as HistoGPA.
For the generative methods STFlow and Stem, predictions are
obtained with the sampling settings of their official
implementations.
\FloatBarrier
\section{Additional Marker-Gene Result}
\label{sup:marker_results}

\begin{figure}[htbp]
    \centering
    \includegraphics[width=\linewidth]{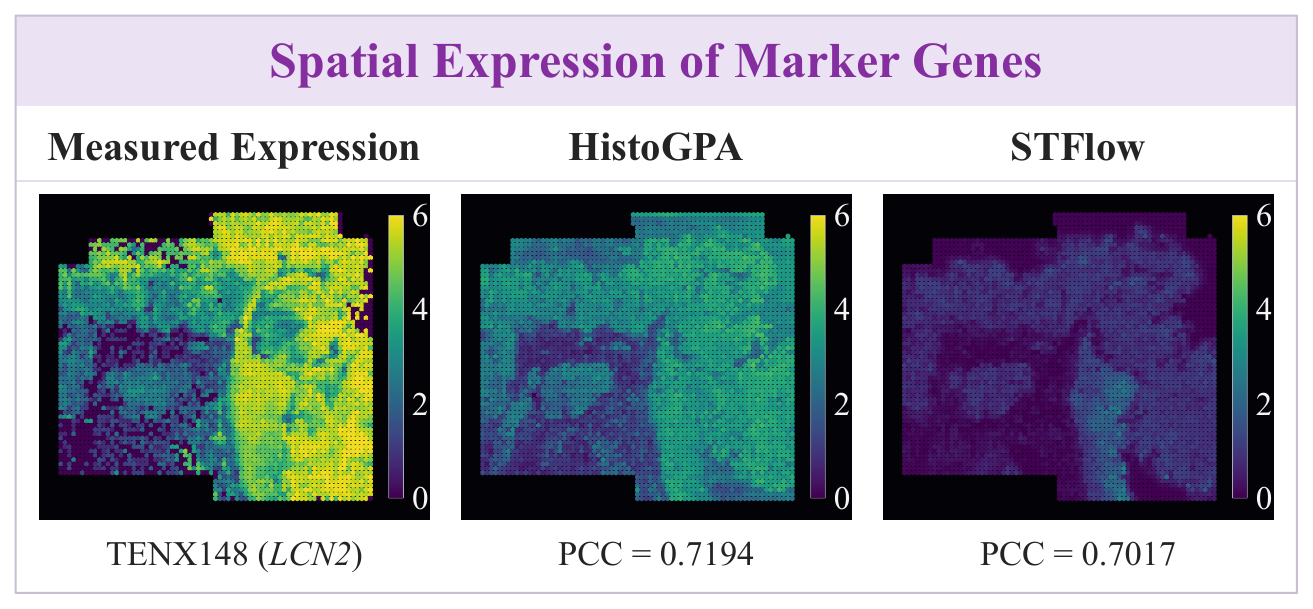}
 \caption{
Additional spatial expression comparison for LCN2 on
TENX148, a held-out COAD test slide.
Columns show Measured Expression, HistoGPA, and STFlow
from left to right.
For each method, the displayed prediction is obtained
from the run whose validation loss is closest to the
median across the 20 independent runs.
The same expression normalization and 0--6 color scale
as in Fig.~2 are used.
PCC across valid tissue locations is reported beneath
each prediction.
}
    \label{sup:fig_marker}
\end{figure}

Section~4.4 presents spatial
expression maps for selected held-out test slides.
Figure~\ref{sup:fig_marker} provides an additional visualization of LCN2
on TENX148, a held-out COAD test slide of split~0,
complementing the examples shown in Fig.~2.
LCN2 and GPRC5A were selected from prior literature before
any predictions were inspected.

As in Sec.~4.4, the displayed
prediction for each method is obtained from the run whose
validation loss is closest to the median across the 20
independent runs, without using test-set prediction
performance for run selection.
On TENX148, HistoGPA obtains a PCC of 0.7194, compared
with 0.7017 for STFlow, showing a smaller but directionally
consistent improvement.

\FloatBarrier
\section{Clustering Evaluation Details}
\label{sup:clustering}

\subsection{Clustering Protocol}
\label{sup:clustering_protocol}

For each held-out test slide, measured and predicted expression
profiles are clustered independently using the complete
1,500-gene target panel.
Genes with zero variance within the corresponding expression
matrix are first removed.
The remaining gene dimensions are standardized to zero mean
and unit variance, with scaled values clipped at 10.
We then apply principal component analysis (PCA), retaining
up to 50 principal components, and construct a
15-nearest-neighbor graph in the resulting PCA space.
Leiden clustering is subsequently performed with resolution
$0.5$ and random seed $0$.
The same preprocessing and clustering settings are applied
independently to measured and predicted expression.
The number of clusters is neither fixed nor transferred
from the measured expression; instead, it is determined
independently by Leiden clustering for each expression matrix.
Spatial coordinates and histological information are not
used during clustering and are introduced only when
visualizing the resulting partitions.

We use adjusted Rand index (ARI) as the primary
partition-agreement metric because it is symmetric,
adjusted for chance agreement, and invariant to
cluster-label permutations.
Adjusted mutual information (AMI), normalized mutual
information (NMI), and homogeneity are additionally
computed as complementary metrics.
AMI and NMI are computed using arithmetic averaging.
For homogeneity, the partition obtained from measured
expression is treated as the reference labeling.

For each model and cancer-specific task, 20 independently
trained runs are evaluated under each of the two predefined
split configurations.
For a given random seed, slide-level clustering metrics
from both split configurations are pooled and averaged
across all held-out test slides within each cancer type.
Since the two split configurations contain different
numbers of held-out slides, scores are averaged across
individual slides rather than by equally averaging the
two split-level means.
The resulting cancer-level scores are then averaged
uniformly across the ten cancer types to obtain one
run-level macro-average.
We report the mean and standard deviation of these
20 run-level macro-averaged scores.
No model run is selected or excluded according to
test-set clustering performance.

For the qualitative maps in Figure~3,
we select, for each method and corresponding cancer-specific
task, the run whose validation loss is closest to the median
across the 20 independent runs.
This selection is independent of all test-set clustering
metrics.
To facilitate visual comparison, predicted cluster IDs
are matched to measured-expression cluster IDs by
maximizing the overlap of paired spatial locations using
the Hungarian algorithm.
This relabeling is applied only after all quantitative
metrics have been computed and therefore has no effect
on the reported scores.
Cluster IDs are otherwise arbitrary and should not be
interpreted as predefined biological cell types or
tissue domains.
\FloatBarrier
\section{Additional Ablation Details}
\label{sup:ablation}

\subsection{Slide-Context Controls}
\label{sup:context_controls}

The slide-context controls in Table~\ref{tab:ablation}
replace the slide-level representation $u$ with $\bar{u}$, the
element-wise mean of $u$ over the training slides used to train
the corresponding run.
Validation and test slides are excluded from this mean, so
$\bar{u}$ contains no information about any held-out slide.

The three mean-context variants have the same architecture and
number of trainable parameters as HistoGPA.
They differ only in the input to each pathway:
Mean context (both) supplies $\bar{u}$ to GCM and to the
conditioning network of GPA;
Mean context in GPA supplies $\bar{u}$ to the conditioning
network of GPA and $u$ to GCM;
Mean context in GCM supplies $\bar{u}$ to GCM and $u$ to the
conditioning network of GPA.
For each run, $\bar{u}$ is fixed and used during training,
validation, and testing.
Each variant is trained independently with the optimization
settings, model-selection criterion, and random seeds of
Sec.~\ref{sup:optimization}.

For Test-time mean, we use the trained HistoGPA checkpoints
reported in Table~\ref{tab:main_results} without retraining or
fine-tuning, and replace $u$ with the same $\bar{u}$ in both
pathways at test time.
The patch-level representations, spatial coordinates, pretrained
gene embeddings, and all model parameters remain unchanged.

\subsection{Per-Cancer Results}
\label{sup:ablation_results}

Tables~\ref{sup:tab_ablation_main} and~\ref{sup:tab_ablation_context}
report per-cancer results for all variants in Table~\ref{tab:ablation},
using the evaluation protocol of Sec.~4.1.

\begin{table*}[t]
\centering
\caption{
Per-cancer results of the module ablation and gene-prior controls
in Table~\ref{tab:ablation}, reported as gene-wise PCC.
Base removes both GCM and GPA;
w/o GPA retains only GCM;
w/o GCM retains context-conditioned GPA without GCM.
Static disables slide-dependent gene-memory modulation
and directly uses the projected pretrained embeddings
in cross-attention, while keeping GCM disabled.
Frozen and Trainable replace the pretrained gene embeddings
with frozen and trainable randomly initialized tokens,
respectively, while retaining GCM and context-conditioned GPA.
Results are mean $\pm$ SD across 20 independent runs, and
Macro Avg.\ is the unweighted mean across the ten cancer types,
reported as mean $\pm$ SD of the per-run macro averages.
Bold indicates the best mean result in each row.
}
\label{sup:tab_ablation_main}
\footnotesize
\setlength{\tabcolsep}{2pt}
\renewcommand{\arraystretch}{1.08}
\newcommand{\msd}[2]{\ensuremath{#1\,{\scriptstyle\pm #2}}}
\newcommand{\bmsd}[2]{\ensuremath{\mathbf{#1}\,{\scriptstyle\pm #2}}}
\begin{tabular*}{\textwidth}{@{\extracolsep{\fill}}lccccccc@{}}
\toprule
& \multicolumn{3}{c}{\textbf{Module Ablation}}
& \multicolumn{3}{c}{\textbf{Gene-Prior Controls}}
& \textbf{Full Model} \\
\cmidrule(lr){2-4}\cmidrule(lr){5-7}\cmidrule(l){8-8}
Cancer Type & Base & w/o GPA & w/o GCM & Static & Frozen & Trainable & HistoGPA \\
\midrule
CCRCC & \msd{0.114}{0.011} & \msd{0.112}{0.008} & \msd{0.095}{0.007} & \msd{0.116}{0.003} & \msd{0.132}{0.015} & \msd{0.133}{0.016} & \bmsd{0.144}{0.004} \\
COAD & \msd{0.235}{0.008} & \msd{0.273}{0.017} & \msd{0.299}{0.016} & \msd{0.258}{0.015} & \msd{0.312}{0.023} & \msd{0.316}{0.008} & \bmsd{0.332}{0.020} \\
HCC & \msd{0.080}{0.016} & \msd{0.085}{0.023} & \msd{0.091}{0.011} & \msd{0.074}{0.017} & \msd{0.129}{0.009} & \msd{0.131}{0.009} & \bmsd{0.137}{0.014} \\
IDC & \msd{0.401}{0.005} & \msd{0.419}{0.021} & \msd{0.435}{0.019} & \msd{0.422}{0.023} & \msd{0.451}{0.012} & \msd{0.449}{0.011} & \bmsd{0.481}{0.015} \\
LUNG & \msd{0.296}{0.028} & \msd{0.407}{0.043} & \msd{0.396}{0.025} & \msd{0.306}{0.021} & \msd{0.403}{0.039} & \msd{0.413}{0.032} & \bmsd{0.429}{0.030} \\
LYMPH-IDC & \msd{0.191}{0.014} & \msd{0.187}{0.029} & \msd{0.199}{0.013} & \msd{0.198}{0.010} & \msd{0.208}{0.024} & \msd{0.209}{0.021} & \bmsd{0.228}{0.018} \\
PAAD & \msd{0.241}{0.006} & \msd{0.318}{0.049} & \msd{0.319}{0.017} & \msd{0.239}{0.025} & \msd{0.382}{0.017} & \msd{0.384}{0.013} & \bmsd{0.411}{0.020} \\
PRAD & \msd{0.143}{0.009} & \msd{0.149}{0.026} & \msd{0.157}{0.023} & \msd{0.151}{0.017} & \msd{0.180}{0.018} & \msd{0.185}{0.009} & \bmsd{0.194}{0.007} \\
READ & \msd{0.184}{0.021} & \msd{0.176}{0.016} & \msd{0.181}{0.014} & \msd{0.157}{0.012} & \msd{0.185}{0.011} & \msd{0.189}{0.005} & \bmsd{0.206}{0.017} \\
SKCM & \msd{0.374}{0.007} & \msd{0.370}{0.017} & \msd{0.390}{0.012} & \msd{0.375}{0.008} & \msd{0.403}{0.014} & \msd{0.407}{0.012} & \bmsd{0.437}{0.019} \\
\midrule
Macro Avg. & \msd{0.226}{0.008} & \msd{0.250}{0.014} & \msd{0.256}{0.008} & \msd{0.230}{0.012} & \msd{0.279}{0.011} & \msd{0.282}{0.009} & \bmsd{0.300}{0.006} \\
\bottomrule
\end{tabular*}
\end{table*}

\begin{table*}[t]
\centering
\caption{
Per-cancer results of the slide-context controls in
Table~\ref{tab:ablation}, reported as gene-wise PCC
(mean $\pm$ SD across 20 independent runs).
Macro Avg.\ is the unweighted mean across the ten cancer types,
reported as mean $\pm$ SD of the per-run macro averages.
$\bar{u}$ is defined in Sec.~\ref{sup:context_controls}.
$^\dagger$Trained HistoGPA models evaluated with $\bar{u}$ in both
pathways, without retraining.
Bold indicates the best mean result in each row.
}
\label{sup:tab_ablation_context}
\footnotesize
\setlength{\tabcolsep}{2pt}
\renewcommand{\arraystretch}{1.08}
\newcommand{\msd}[2]{\ensuremath{#1\,{\scriptstyle\pm #2}}}
\newcommand{\bmsd}[2]{\ensuremath{\mathbf{#1}\,{\scriptstyle\pm #2}}}
\begin{tabular*}{\textwidth}{@{\extracolsep{\fill}}lccccc@{}}
\toprule
Cancer Type & Mean context (both) & Mean context in GPA & Mean context in GCM & Test-time mean$^\dagger$ & HistoGPA \\
\midrule
CCRCC & \msd{0.128}{0.018} & \msd{0.135}{0.019} & \msd{0.132}{0.013} & \msd{0.135}{0.015} & \bmsd{0.144}{0.004} \\
COAD & \msd{0.301}{0.013} & \msd{0.315}{0.016} & \msd{0.307}{0.009} & \msd{0.317}{0.006} & \bmsd{0.332}{0.020} \\
HCC & \msd{0.105}{0.017} & \msd{0.122}{0.011} & \msd{0.122}{0.014} & \msd{0.129}{0.008} & \bmsd{0.137}{0.014} \\
IDC & \msd{0.433}{0.009} & \msd{0.454}{0.012} & \msd{0.444}{0.015} & \msd{0.446}{0.018} & \bmsd{0.481}{0.015} \\
LUNG & \msd{0.360}{0.010} & \msd{0.398}{0.021} & \msd{0.387}{0.036} & \msd{0.382}{0.032} & \bmsd{0.429}{0.030} \\
LYMPH-IDC & \msd{0.194}{0.014} & \msd{0.204}{0.024} & \msd{0.207}{0.043} & \msd{0.198}{0.017} & \bmsd{0.228}{0.018} \\
PAAD & \msd{0.289}{0.007} & \msd{0.378}{0.020} & \msd{0.366}{0.022} & \msd{0.334}{0.007} & \bmsd{0.411}{0.020} \\
PRAD & \msd{0.148}{0.019} & \msd{0.174}{0.017} & \msd{0.178}{0.012} & \msd{0.179}{0.008} & \bmsd{0.194}{0.007} \\
READ & \msd{0.167}{0.006} & \msd{0.183}{0.009} & \msd{0.180}{0.005} & \msd{0.182}{0.011} & \bmsd{0.206}{0.017} \\
SKCM & \msd{0.373}{0.004} & \msd{0.379}{0.011} & \msd{0.381}{0.007} & \msd{0.403}{0.005} & \bmsd{0.437}{0.019} \\
\midrule
Macro Avg. & \msd{0.250}{0.008} & \msd{0.274}{0.009} & \msd{0.270}{0.011} & \msd{0.271}{0.005} & \bmsd{0.300}{0.006} \\
\bottomrule
\end{tabular*}
\end{table*}

\end{document}